\pdfoutput=1

\documentclass[11pt]{article}

\usepackage[final]{acl}

\usepackage{times}
\usepackage{latexsym}

\usepackage[T1]{fontenc}

\usepackage[utf8]{inputenc}

\usepackage{microtype}

\usepackage{inconsolata}

\usepackage{graphicx}
\usepackage{amsmath}
\usepackage{color}
\usepackage{tabularx}
\usepackage{booktabs}
\usepackage{array}
\newcolumntype{Y}{>{\centering\arraybackslash}X}
\usepackage{multirow}
\usepackage{latexsym}
\usepackage{amssymb}
\usepackage{amsmath}
\usepackage{amsthm}
\usepackage{enumitem}
\usepackage{multirow}
\usepackage{float}
\usepackage{stfloats}
\usepackage{nicematrix}
\usepackage{listings}

\title{ES-Mem: Event Segmentation-Based Memory for Long-Term 

Dialogue Agents}

\author{
    \textbf{Huhai Zou}$^{1}$, 
    \textbf{Tianhao Sun}$^{1\dagger}$, 
    \textbf{Chuanjiang He}$^{1}$, 
    \textbf{Yu Tian}$^{2}$, 
    \textbf{Zhenyang Li}$^{3}$, \\
    \textbf{Li Jin}$^{4}$, 
    \textbf{Nayu Liu}$^{5}$, 
    \textbf{Jiang Zhong}$^{1}$, 
    \textbf{Kaiwen Wei}$^{1\dagger}$ \\
    $^{1}$College of Computer Science, Chongqing University \quad
    $^{2}$Tsinghua University \\
    $^{3}$Hong Kong Generative AI Research \& Development Center, HKUST\\
    $^{4}$Aerospace Information Research Institute, Chinese Academy of Sciences \\
    $^{5}$School of Computer Science and Technology, Tiangong University \\
    {\tt \{sthing, weikaiwen\}@cqu.edu.cn}
}

\begin{document}
\maketitle
\begin{abstract}

Memory is critical for dialogue agents to maintain coherence and enable continuous adaptation in long-term interactions. While existing memory mechanisms offer basic storage and retrieval capabilities, they are hindered by two primary limitations: (1) rigid memory granularity often disrupts semantic integrity, resulting in fragmented and incoherent memory units; (2) prevalent flat retrieval paradigms rely solely on surface-level semantic similarity, neglecting the structural cues of discourse required to navigate and locate specific episodic contexts.
To mitigate these limitations, drawing inspiration from Event Segmentation Theory, we propose ES-Mem, a framework incorporating two core components: (1) a dynamic event segmentation module that partitions long-term interactions into semantically coherent events with distinct boundaries; (2) a hierarchical memory architecture that constructs multi-layered memories and leverages boundary semantics to anchor specific episodic memory for precise context localization. Evaluations on two memory benchmarks demonstrate that ES-Mem yields consistent performance gains over baseline methods. Furthermore, the proposed event segmentation module exhibits robust applicability on dialogue segmentation datasets. 

\end{abstract}

\section{Introduction}

Large Language Models (LLMs) have fundamentally transformed the landscape of dialogue agents, offering unprecedented generative capabilities\citep{llm-persona, evaluating-agents}. To sustain coherent long-term interactions beyond the fixed context window inherent to self-attention mechanisms \citep{2}, agents require a dedicated memory module \citep{memory-matters, human-ai-memory, rethinking-memory}. From the perspectives of cognitive psychology, self-evolution, and agent applications \citep{memory-servey}, memory serves as the pivotal infrastructure bridging conversational context with current reasoning to foster intelligent and personalized evolution.

Existing methods can be primarily categorized by their memory storage forms. Some studies utilize textual vector-based representations \citep{MemGPT, Memorybank, hello-again, memoryos}, where dialogue history is maintained as raw segments or summaries to support semantic retrieval and context retention. Other approaches advocate for structured graph-based forms \citep{zep, lifelong, A-mem}, organizing information into knowledge graphs to explicitly model entity relationships and enhance logical connectivity.

\begin{figure}[t]
    \centering
    \includegraphics[width=\linewidth]{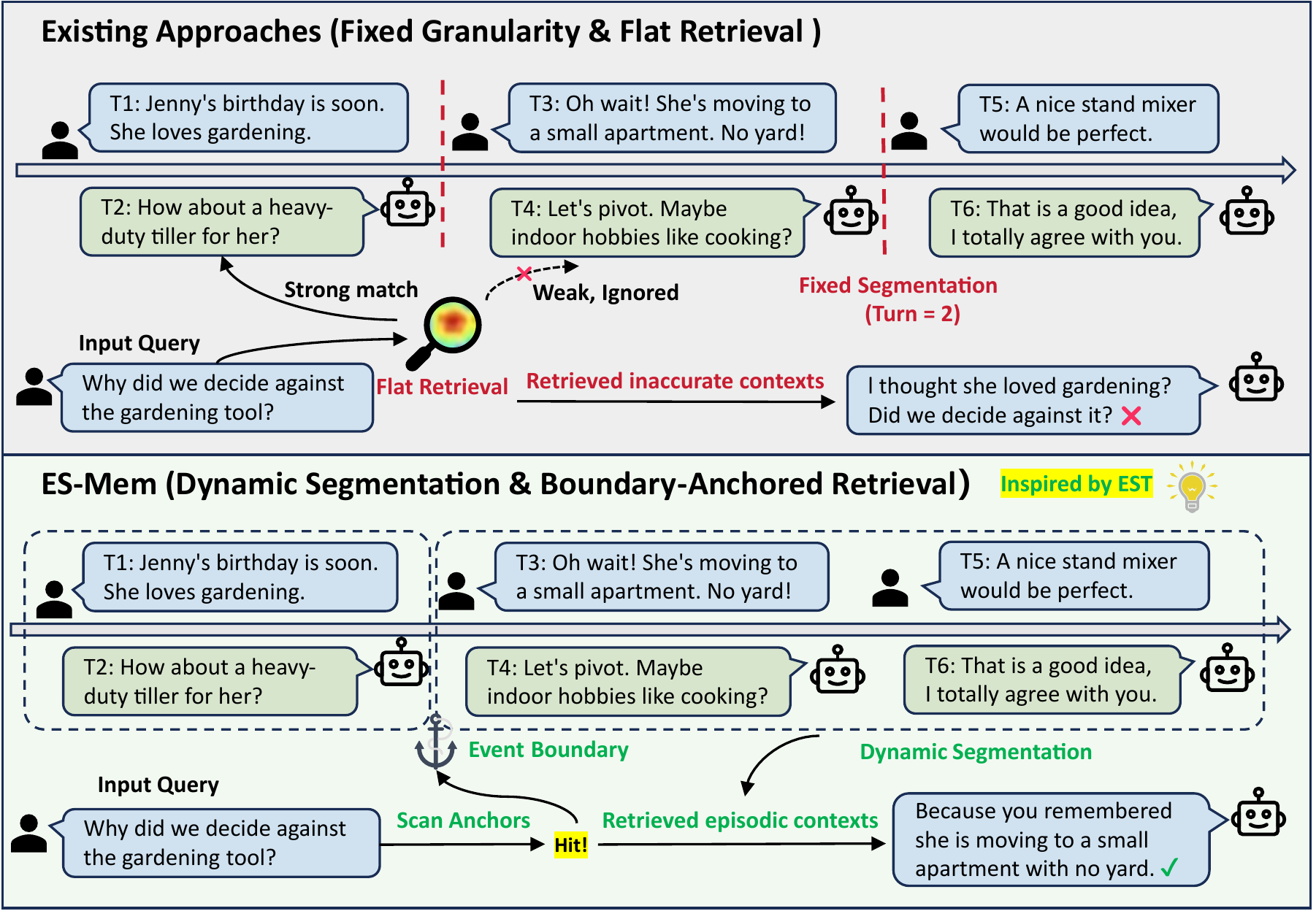}
    \caption{Comparison between existing memory methods and ES-Mem. Existing methods with fixed granularity often sever semantic dependencies, causing flat retrieval to recall inaccurate contexts. Conversely, ES-Mem integrates Event Segmentation Theory (EST) to dynamically structure memory. By using boundary anchors, it precisely locates episodic contexts.}
    \label{fig:existing memory}
\end{figure}

While existing memory methods support basic storage and retrieval functionalities, as illustrated in Figure \ref{fig:existing memory}, they remain constrained by two critical limitations:
\textbf{(1) Memory granularity lacks semantic adaptability.} Prevalent approaches \citep{Memorybank, MemGPT, self-controlled, A-mem} typically utilize fixed dialogue turns as the unit of storage. This rigid granularity disrupts semantic continuity, resulting in fragmented memory units.
\textbf{(2) Retrieval paradigms lack discourse-structural awareness.} Current memory methods \citep{MemGPT, Memorybank, memoryos} predominantly employ flat retrieval strategies that treat memory units as isolated vectors, neglecting the structural cues of discourse. This limitation prevents agents from utilizing the logical boundaries of discourse to accurately locate specific episodic contexts.

To mitigate these limitations, we draw inspiration from Event Segmentation Theory (EST) \citep{event-segment}. This cognitive theory suggests that humans do not perceive experience as a continuous stream, but rather parse it into discrete, meaningful units delimited by event boundaries. Crucially, these boundaries elicit heightened attention and serve as cognitive anchors that compress complex sequential information into structured indices, thereby facilitating efficient long-term memory storage and retrieval.

Building upon these cognitive principles, we propose the Event Segmentation-based Memory framework (ES-Mem). This framework implements the theory through two key innovations. First, \textbf{the dynamic event segmentation module} employs a two-stage approach based on topical coherence and intent transition probabilities to dynamically partition long-term dialogues into semantically coherent segments, effectively alleviating the fragmentation induced by fixed memory granularity. Second, \textbf{the hierarchical memory architecture} establishes a multi-layered structure comprising refined boundaries, event summaries, and raw context. This component adopts a human-like recall strategy that utilizes event boundaries as cognitive anchors to locate episodic memory intervals, followed by fine-grained reranking to retrieve precise context details, thereby addressing the cognitive deficiency of flat retrieval paradigms.

We evaluate ES-Mem on the LoCoMo \citep{Locomo} and LongMemEval-S \citep{longmemeval} benchmarks to validate its effectiveness. Experimental results demonstrate that ES-Mem consistently outperforms baseline methods, confirming that coherent event-level memory and hierarchical retrieval strategy effectively mitigate the noise and semantic fragmentation typical of flat retrieval paradigms. Additionally, the robust applicability of our dynamic event segmentation module is evidenced by its strong performance across the DialSeg711 \citep{dilseg711}, TIAGE \citep{tiage}, and SuperDialSeg \citep{superdialseg} datasets. 
The primary contributions of our work are summarized as follows:

\begin{itemize}
    \item We propose ES-Mem, a novel cognitive-inspired memory framework grounded in Event Segmentation Theory. By shifting memory granularity from rigid turns to dynamic events, ES-Mem addresses the semantic fragmentation inherent in existing approaches and ensures the preservation of discourse integrity.
    
    \item We implement a dynamic segmentation module that partitions continuous dialogue streams based on topical coherence and intent shifts. This drives a hierarchical memory architecture comprising multi-layered storage, which enables a boundary-anchored strategy for precise context localization.
    
    \item We systematically evaluate ES-Mem’s performance on two long-term memory benchmarks. Empirical results demonstrate that ES-Mem consistently outperforms memory baselines. Additionally, our event segmentation module exhibits robust adaptability on dialogue segmentation tasks in small model scenarios.
\end{itemize}

\section{Related Works}
\subsection{Memory for Dialogue Agents}
Existing memory strategies can be broadly categorized into textual chunk-based and knowledge graph-based paradigms. 
Textual approaches typically rely on context chunks or summaries to optimize storage and retrieval. For instance, inspired by operating systems, MemGPT \citep{MemGPT} and MemoryOS \citep{memoryos} are designed to manage context window limitations through an OS-like memory architecture. And MemoryBank \citep{Memorybank} incorporates the Ebbinghaus forgetting curve to model the natural decay of memory strength. A-Mem \citep{A-mem} represents each memory as an atomic note and dynamically establishes and updates links among notes.
Graph-based methods explicitly model structural dependencies. HippoRAG \citep{HipperRAG-1, HipperRAG-2} leverages knowledge graphs to facilitate multi-hop reasoning. To address temporal dynamics, Zep \citep{zep} and MemoTime \citep{memotime} utilize temporal knowledge graphs to track evolving entity states. SGMem \citep{sgmem} organizes memory via sentence-level graphs to retain structural logic alongside textual content.
While prior research has advanced memory modeling, existing methods remain constrained by the rigid memory granularity. To address this, we propose ES-Mem, a framework that employs dynamic event segmentation to preserve semantic integrity.

\subsection{Dynamic Memory Granularity}\label{2.2}
The granularity of memory significantly impacts agent performance, yet fixed approaches (e.g., turn or session-level) often suffer from limited retrieval accuracy and semantic completeness \citep{secom}. Consequently, recent research has shifted towards dynamic partitioning strategies that align with dialogue semantics. 
SeCom \citep{secom} establishes memory units at the granularity of dialogue segments, utilizing structured prompts to directly generate topically coherent chunks.
RMM \citep{rmm} uses forward-looking reflection to dynamically split and integrate dialogue history by semantically coherent topics. 
Prior works Nemori \citep{nemori} and EM-LLM \citep{em-llm} also draw inspiration from Event Segmentation Theory to structure memory. Nemori employs an LLM-based detector with boundary and representation alignment to segment continuous dialogue streams. EM-LLM determines initial boundaries via Bayesian surprise and refines them using graph-theoretic metrics. 
However, these approaches primarily utilize EST for defining storage granularity, neglecting the explicit modeling of boundary semantics. ES-Mem advances this by repurposing boundaries as informational indices that serve as cognitive anchors to guide a structure-aware, coarse-to-fine retrieval process. Furthermore, unlike methods that rely heavily on the extensive reasoning of large-scale LLMs, ES-Mem adopts a lightweight segmentation strategy to ensure resource adaptability.

\begin{figure*}[t]
    \centering
    \includegraphics[width=\linewidth]{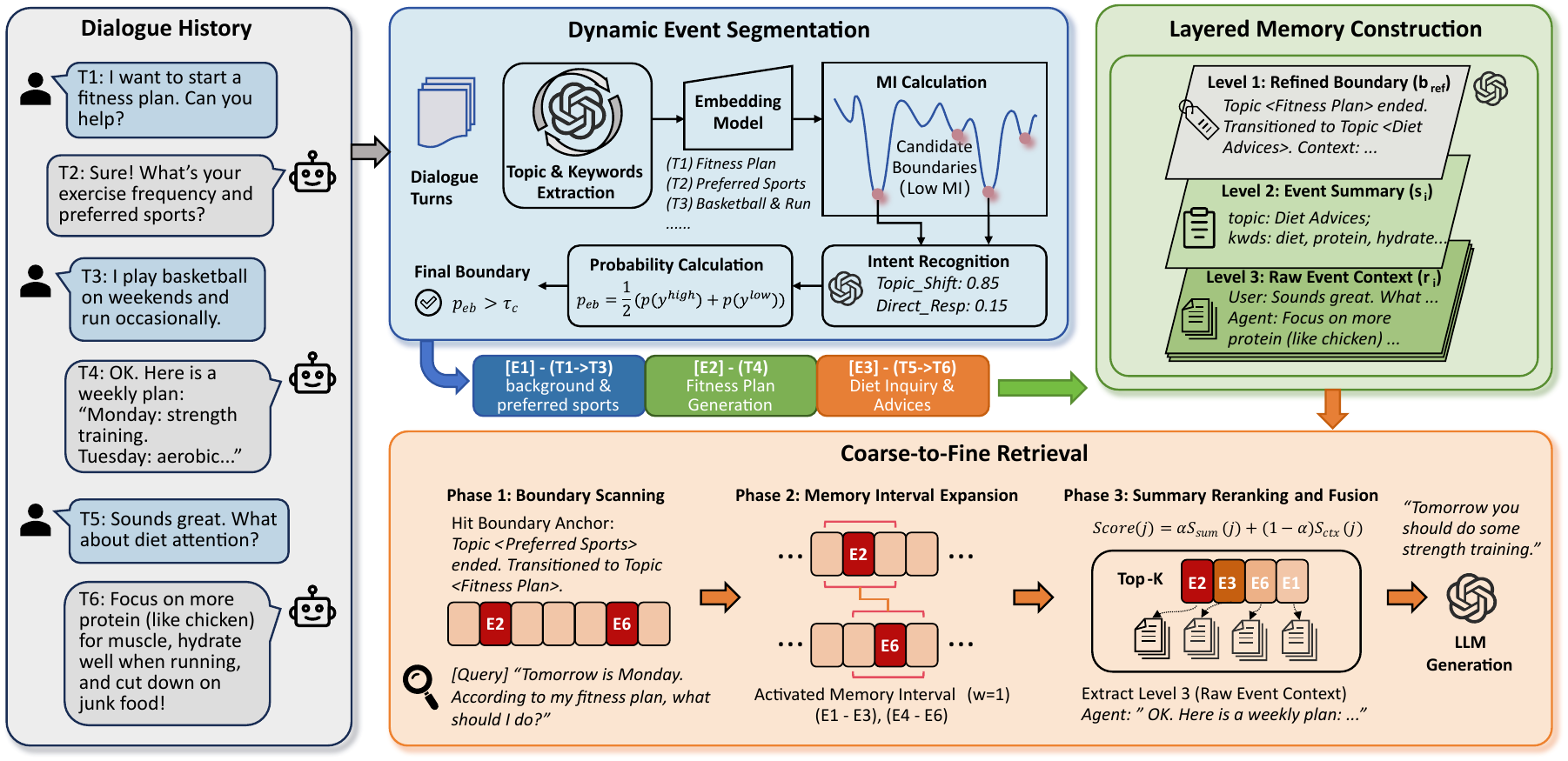}
    \caption{Overview of ES-Mem. The framework
    consists of three modules: Dynamic Event Segmentation partitions the dialogue stream into semantic events; Layered Memory Construction builds multi-level storage with boundary anchors; and Coarse-to-Fine Retrieval utilizes these anchors to precisely locate and expand relevant contexts.}
    \label{fig:ES-Mem}
\end{figure*}

\section{Methodology}

Analogous to the cognitive mechanisms of human memory, we design the ES-Mem framework as illustrated in Figure \ref{fig:ES-Mem}. It comprises two core components: a \textit{Dynamic Event Segmentation} model and a \textit{Hierarchical Memory Architecture}. Specifically, the latter is composed of \textit{Layered Memory Construction} and a \textit{Coarse-to-Fine Retrieval} strategy.

\subsection{Dynamic Event Segmentation}\label{3.1}
We define a dialogue event as a temporal segment centered on a stable topic or intent, characterized by high internal semantic homogeneity and bounded by distinct shifts in topic, task phases, or interaction patterns. To effectively capture these boundaries, we propose a two-stage segmentation strategy that integrates topic coherence detection with intent-aware boundary refinement.

\paragraph{Stage 1: Topic Coherence Detection.}
Given a dialogue session $\mathcal{U} = \{u_1, u_2, \ldots, u_T\}$ consisting of $T$ turns, we first employ an LLM to extract a concise topic representation for each turn $u_t$. To mitigate semantic drift and ensure temporal continuity, we adopt a recurrent extraction scheme: the LLM is conditioned on the concatenation of the previous turn's topic output and the current turn's content. This generates a smoothed topic evolution trajectory $h_t$.
We then utilize an embedding model to map these textual topics into a sequence of normalized vectors $\mathbf{H} = \{ \mathbf{h}_1, \mathbf{h}_2, \dots, \mathbf{h}_T \}$.

To quantify the semantic coupling between adjacent turns, we assume the dimensions of the topic vectors $\mathbf{h}_t$ and $\mathbf{h}_{t+1}$ follow a joint Gaussian distribution. Let random variables $X = \mathbf{h}_t$ and $Y = \mathbf{h}_{t+1}$. We compute the Pearson correlation coefficient $\rho_t$ across their dimensions:
\begin{equation}\label{eq:pearson}
\rho_t = \frac{\mathrm{Cov}(X,Y)}{\sqrt{\mathrm{Var}(X)\mathrm{Var}(Y)}}
\end{equation}

Under the Gaussian assumption, the Mutual Information (MI) between $X$ and $Y$ can be derived as:
\begin{equation}\label{eq:mi}
    I_t = -\frac{1}{2}\log\bigl(1 - \rho_t^2\bigr)
\end{equation}
Here, $I_t$ serves as a proxy for topical consistency. A sharp decline in $I_t$ indicates a rupture in semantic coupling, signaling a potential event boundary. We compute the global MI sequence for the session and identify a set of candidate boundaries $\mathcal{C}$ by selecting time steps where $I_t$ falls below a dynamic quantile threshold $q$ (e.g., the bottom 35\%): $\mathcal{C} = \{t \mid I_t \le \mathrm{Quantile}_q(\mathbf{I})\}$.

\paragraph{Stage 2: Intent-Aware Boundary Refinement.}
The candidate set $\mathcal{C}$ derived from statistical metrics may contain noise. Therefore, we introduce a second stage of intent-level verification using an LLM to align the segmentation with cognitive event theory.
For each candidate boundary $t \in \mathcal{C}$, we construct local context windows comprising the preceding and succeeding $L$ turns: $\mathcal{L}_t = \{u_{t-L+1}, \ldots, u_t\}$ and $\mathcal{R}_t = \{u_{t+1}, \ldots, u_{t+L}\}$. This localized context enables the LLM to discern subtle shifts in intent without ingesting the entire history.

Drawing inspiration from \citet{def-dts}, we define a label set partitioned into boundary-positive labels $\mathcal{Y}_{\text{shift}}$ (e.g., \textsc{Topic\_Shift}, \textsc{Topic\_Intro}) and boundary-negative labels $\mathcal{Y}_{\text{cont}}$ (e.g., \textsc{Detail\_Elaborate}, \textsc{Direct\_Resp}). The taxonomy is detailed in Appendix Figure \ref{tab:intent-label}.
The LLM analyzes the transition from $\mathcal{L}_t$ to $\mathcal{R}_t$ and outputs the most probable labels along with their confidence scores $c_t \in [0,1]$. We formulate the probability of $t$ being a valid event boundary, $p_{\mathrm{eb}}(t)$, by aggregating the model's confidence in shift versus continuation:
\begin{equation}\label{eq:prob_map}
p(y_t) =
\begin{cases}
c_t,      & \text{if } y_t \in \mathcal{Y}_{\text{shift}}\\
1 - c_t,  & \text{if } y_t \in \mathcal{Y}_{\text{cont}}
\end{cases}
\end{equation}

To ensure robustness, we average the probabilities derived from the highest confidence prediction ($p_h$) and the lowest confidence prediction ($p_l$):
\begin{equation}
    p_{\mathrm{eb}}(t) = \frac{1}{2}\bigl(p(y_{t}^{\text{high}}) + p(y_{t}^{\text{low}})\bigr)
\end{equation}

Finally, we apply a confidence threshold $\tau_c$. Positions satisfying $p_{\mathrm{eb}}(t) \ge \tau_c$ constitute the final boundary set $\mathcal{B}$, partitioning the dialogue into semantically coherent and structurally distinct event units.

\subsection{Layered Memory Construction}
Unlike previous approaches that rely on flat storage or complex graph structures, we construct a layered memory comprising three levels: \textit{Level 1: Refined Boundaries}, \textit{Level 2: Event Summaries}, and \textit{Level 3: Raw Context}.
After dynamic event segmentation, the memory repository is defined as $\mathcal{M} = \{M_1, M_2, \ldots, M_n\}$. Each memory unit $M_i$ is formally represented as:
\begin{equation}
    M_i = (b_i^{\text{ref}}, s_i, r_i, t_i), \quad \forall M_i \in \mathcal{M}
\end{equation}
where $t_i$ denotes the timestamp of the event.

\paragraph{Level 1: Refined Event Boundaries ($b_i^{\text{ref}}$).}
Existing memory mechanisms predominantly focus on the internal semantic content of individual events, typically neglecting the explicit modeling of boundaries or state transitions. This oversight results in a loss of critical structural cues that serve as natural cognitive anchors for retrieval. To address this gap, we propose a \textit{Refined Boundary Representation} that explicitly models the state transition between events.
Specifically, we employ an LLM to generate a textual description of the transition from event $M_{i-1}$ to $M_i$, leveraging their summaries ($s_{i-1}, s_i$) and the raw boundary context ($c_i^{\text{raw}}$, consisting of the last $L$ turns of $M_{i-1}$ and the first $L$ turns of $M_i$). The generation process is formalized as:
\begin{equation}
    b_i^{\text{ref}} = \text{LLM}_{\text{gen}}(s_{i-1}, s_i, c_i^{\text{raw}})
\end{equation}

The resulting $b_i^{\text{ref}}$ follows a structured format (e.g., ``\textit{Topic A ended. Transitioned to Topic B. Context: ...}''), serving as a high-level cognitive anchor.

\paragraph{Level 2: Event Summaries ($s_i$).}
Generated during the dynamic event segmentation phase, the event summary $s_i$ is composed of the core topics and keywords extracted from each dialogue turn within the event. This dense semantic representation serves as the index for fine-grained content matching, bridging the gap between high-level conceptual anchors and low-level verbatim details.

\paragraph{Level 3: Raw Context ($r_i$).}
This level stores the verbatim dialogue records, preserving the complete temporal sequence and semantic fidelity of the original interaction. While Level 1 and Level 2 are used for efficient indexing and retrieval, $r_i$ serves as the \textit{ground truth} source provided to the dialogue agent for final response generation, ensuring that the generated answers are factually accurate and contextually consistent.

\subsection{Coarse-to-Fine Retrieval}
Prevalent flat retrieval paradigms often struggle to locate specific episodic contexts within massive historical data due to the lack of structural guidance. To mitigate this, we propose a coarse-to-fine retrieval strategy that mimics the human cognitive process of ``scanning anchors $\to$ unpacking details.'' This process consists of three stages.

\paragraph{Stage 1: Boundary Scanning (Anchor Retrieval).}
Given a query $q$, we first encode it into vector $e_q$. Instead of matching the query against all event details immediately, we explicitly calculate the similarity between the query and the Refined Boundaries (Level 1) to identify potential state transitions or topic shifts relevant to the query:
\begin{equation}
    \mathrm{sim}_{\mathrm{bnd}}(q,i) = \cos(e_q, e_i^{\mathrm{bnd}}) = \frac{e_q \cdot e_i^{\mathrm{bnd}}}{\lVert e_q \rVert \lVert e_i^{\mathrm{bnd}} \rVert}
\end{equation}

We select the top-$k$ events with the highest boundary similarity scores to form the anchor set $\mathcal{A} = \{a_1, a_2, \ldots, a_k\}$. These anchors represent the moments where the conversation context aligns most closely with the user's inquiry intent.

\paragraph{Stage 2: Activated Memory Interval Expansion.}
Relying solely on single time-point boundaries may miss the full context of an evolving discussion. Therefore, we treat each anchor $a \in \mathcal{A}$ as a center point and expand a temporal window of size $w$ to define an \textit{Activated Memory Interval}. The candidate set $\mathcal{C}$ is formed by the union of these intervals:
\begin{equation}
    \mathcal{C} = \bigcup_{a \in \mathcal{A}} \{j \mid |j - a| \le w\}
\end{equation}

This step significantly filters the search space, focusing only on logically relevant intervals. Importantly, each candidate event $j \in \mathcal{C}$ inherits a \textit{context score} $S_{\text{ctx}}(j)$ from its generating anchor, defined as the maximum boundary score within its influential window:
\begin{equation}
    S_{\text{ctx}}(j) = \max_{\{a \in \mathcal{A} \mid j \in \text{window}(a)\}} \mathrm{sim}_{\mathrm{bnd}}(q, a)
\end{equation}

\paragraph{Stage 3: Summary Reranking and Fusion.}
In the final stage, we perform fine-grained matching within the candidate set $\mathcal{C}$ using Level 2 Event Summaries. We compute the content similarity $\mathrm{sim}_{\mathrm{sum}}(q,j)$ for each $j \in \mathcal{C}$. The final ranking score is a fusion of the specific content match and the inherited context relevance:
\begin{equation}
    Score(j) = \alpha S_{\text{sum}}(j) + (1-\alpha) S_{\text{ctx}}(j)
\end{equation}
where $\alpha$ is a balancing hyperparameter and $S_{\text{sum}}(j)$ denotes $\mathrm{sim}_{\mathrm{sum}}(q,j)$. We select the top-$K$ events with the highest final scores and retrieve their raw records $r_j$ (Level 3) to construct the context for the LLM agent response.

\section{Experiments}
\subsection{Experimental Settings}\label{4.1}

\paragraph{Datasets.}
We evaluate ES-Mem on two long-term memory benchmarks. \textbf{LoCoMo} \citep{Locomo} comprises 10 extensive conversations, each averaging approximately 600 turns. It covers single-hop, multi-hop, temporal reasoning, and open-domain question answering. \textbf{LongMemEval-S} \citep{longmemeval} has a larger data scale than LoCoMo, consisting of 500 conversations between users and assistants, and each conversation contains an average of 50 sessions and 110k tokens.

\paragraph{Evaluation Metrics.}
We follow A-Mem \citep{A-mem} to use \textbf{F1} and \textbf{BLEU-1 (B1)} as the evaluation metrics of LoCoMo benchmark. For LongMemEval-S, we adopt the prompt templates from \citet{longmemeval} and use an LLM judge to measure the \textbf{Accuracy (ACC)} of the generated answer by comparing the ground truth.

\paragraph{Baselines.}
We compare ES-Mem against ten baselines of dialogue agent memory: MemGPT \citep{MemGPT}, MemoryBank \citep{Memorybank} A-Mem \citep{A-mem}, MemoryOS \citep{memoryos}, H-Mem \citep{h-mem}, Zep \citep{zep}, Mem0 \citep{mem0}, LangMem\footnote{\label{fn:langmem}\url{https://langchain-ai.github.io/langmem/}}, Nemori \citep{nemori} and LightMem \citep{lightmem}. More details about these baselines can be found in Appendix \ref{sec:baselines}.

\paragraph{Implementation Details.}
We employ three LLMs as backbones, including two small open-source models, Qwen2.5-3B and Llama3.2-3B, and one proprietary model, GPT-4o-mini.
The open-source models perform inference and generation via the Ollama\footnote{ \url{https://ollama.com/}}, while GPT-4o-mini is accessed through an API.
We utilize all-MiniLM-L6-v2\citep{Minilm} model to generate text embeddings and build a Faiss\citep{Faiss} vector index to support efficient similarity retrieval. Further experiments and parameter settings are detailed in Appendix \ref{sec:setup}. And the prompt templates utilized in this study are provided in Appendix \ref{sec:prompt}.

\begin{table*}[htbp]
\centering
\resizebox{\textwidth}{!}{
\setlength{\tabcolsep}{10pt} 
\renewcommand{\arraystretch}{0.92}
\begin{tabular}{cccccccccccc}
\toprule
\multirow{4}{*}{\textbf{LLM}} & \multirow{4}{*}{\textbf{Method}} &
\multicolumn{10}{c}{\textbf{Category}} \\
\cmidrule(lr){3-12}
& & \multicolumn{2}{c}{\textbf{Single Hop}} &
\multicolumn{2}{c}{\textbf{Multi Hop}} &
\multicolumn{2}{c}{\textbf{Temporal}} &
\multicolumn{2}{c}{\textbf{Open Domain}} &
\multicolumn{2}{c}{\textbf{Overall}} \\
\cmidrule(lr){3-4} \cmidrule(lr){5-6} \cmidrule(lr){7-8} \cmidrule(lr){9-10} \cmidrule(lr){11-12}
& & F1 & B1 & F1 & B1 & F1 & B1 & F1 & B1 & F1 & B1 \\
\midrule

\multirow{6}{*}{Qwen2.5-3B}
& MemGPT    & 7.26 & 5.52 & 5.07 & 4.31  & 2.94 & 2.95 & 7.04  & 7.10  & 5.94 & 4.86 \\
& MemoryBank    & 4.11  & 3.32 & 3.60 & 3.39  & 1.72 & 1.97 & 6.63  & 6.58  & 3.67 & 3.25 \\
& A-Mem    & 17.23 & 13.12 & 12.57 & 9.01  & 27.59 & 25.07 & 7.12  & 7.28  & 17.91 & 14.49 \\
& MemoryOS & \underline{26.23} & \underline{22.39} & \textbf{23.26} & \textbf{15.39} & 21.44 & 14.95 & 10.18 & 8.18  & 23.69 & 18.68 \\
& H-Mem    & 24.24 & 19.24 & 18.37 & 12.23 & \underline{31.25} & \textbf{26.36} & \textbf{16.23} & \textbf{13.27} & \underline{24.13} & \underline{19.07} \\
&  \textbf{ES-Mem (Ours)}     & \textbf{35.47} & \textbf{30.54} & \underline{21.68} & \underline{15.18} & \textbf{31.48} & \underline{26.22} & \underline{15.37} & \underline{11.56} & \textbf{30.86} & \textbf{25.65} \\
\midrule

\multirow{6}{*}{Llama3.2-3B}
& MemGPT    & 4.32 & 3.51 & 5.32 & 3.99  & 2.68 & 2.72 & 5.64  & 5.54 & 4.24 & 3.56 \\
& MemoryBank    & 7.61  & 6.03 & 6.19 & 4.47  & 3.49 & 3.13 & 4.07  & 4.57  & 6.27 & 5.04 \\
& A-Mem    & 28.14 & 23.87 & 17.44 & 11.74 & \textbf{26.38} & \underline{19.50} & 12.53 & \underline{11.83} & 24.83 & 19.98 \\
& MemoryOS & 28.15 & 22.91 & 19.99 & \underline{14.50} & 18.04 & 14.25 & 12.37 & 8.60  & 23.56 & 18.67 \\
& H-Mem    & \underline{29.37} & \underline{24.34} & \underline{20.23} & 13.24 & 24.34 & 18.24 & \textbf{17.23} & \textbf{13.27} & \underline{25.89} & \underline{20.35} \\
&  \textbf{ES-Mem (Ours)}     & \textbf{35.18} & \textbf{30.59} & \textbf{26.78} & \textbf{20.53} & \underline{25.83} & \textbf{20.69} & \underline{15.42} & 11.15 & \textbf{30.46} & \textbf{25.48} \\
\midrule

\multirow{9}{*}{GPT-4o-mini}
& MemGPT    & 41.01 & 34.34 & 26.65 & 17.72  & 25.52 & 19.44 & 9.15  & 7.44 & 33.17 & 26.51 \\
& MemoryBank    & 6.61  & 5.16 & 5.00 & 4.77  & 9.68 & 6.99 & 5.56  & 5.94  & 6.89 & 5.52 \\
& A-Mem    & 44.65 & 37.06 & 27.02 & 20.09 & 45.85 & 36.67 & 12.14 & 12.00 & 39.65 & 32.30 \\
& MemoryOS & 48.62 & \underline{42.99} & 35.27 & 25.22 & 41.15 & 30.76 & 20.02 & 16.52 & 42.84 & 35.53 \\
& Zep      & \underline{49.56} & 38.92 & 35.74 & 23.30 & 42.00 & 34.53 & 19.37 & 14.82 & 43.56 & 33.64 \\
& Mem0     & 47.65 & 38.72 & \textbf{38.72} & \underline{27.13} & 48.93 & 40.51 & \underline{28.64} & 21.58 & \underline{45.09} & 35.91 \\
& LangMem  & 40.91 & 33.63 & 35.51 & 26.86 & 30.75 & 25.84 & 26.04 & \textbf{22.32} & 36.87 & 30.06 \\
& Nemori   & 46.33 & 38.52 & 32.36 & 23.36 & \textbf{55.99} & \textbf{48.25} & \textbf{29.19} & \underline{22.31} & 44.72 & 36.75 \\
& LightMem    & 47.64 & 40.91 & 32.11 & 23.79 & \underline{53.79} & \underline{46.96} & 26.14 & 20.19 & 44.73 & \underline{37.74} \\
& \textbf{ES-Mem (Ours)}     & \textbf{50.07} & \textbf{45.21} & \underline{36.52} & \textbf{27.75} & 47.90 & 42.12 & 24.77 & 18.51 & \textbf{45.56} & \textbf{39.70} \\
\bottomrule
\end{tabular}
}
\caption{Results on LoCoMo benchmark across different LLMs and memory methods.
The best performance is marked in \textbf{bold}, and the second best is \underline{underlined}.}
\label{tab:locomo-results}
\end{table*}

\subsection{Main Results}\label{4.2}
\paragraph{Results on LoCoMo.}
As presented in Table \ref{tab:locomo-results}, ES-Mem consistently outperforms baseline methods across all LLM backbones in terms of the overall score. It is worth noting that the performance gains on small-parameter models, such as Qwen2.5-3B, are significantly more pronounced compared to those on GPT-4o-mini. This phenomenon suggests that our memory framework effectively mitigates the cognitive load of context selection, thereby enabling smaller models to transcend their inherent reasoning limitations.
In task-specific evaluations, ES-Mem excels in Single-Hop QA. Utilizing the GPT-4o-mini backbone, our method secures the top position with an F1 score of 50.07. We attribute this precision to our dynamic event segmentation, which ensures the semantic homogeneity of each event unit and facilitates high-fidelity factual matching via hierarchical memories.
Regarding Multi-Hop QA, which demands complex cross-context reasoning, ES-Mem exhibits robust capabilities. It achieves the best F1 score of 26.78 on the Llama3.2-3B backbone and maintains competitive performance on GPT-4o-mini with an F1 score of 36.52. These findings validate the efficacy of our cognitive retrieval mechanism, which effectively reconstructs coherent narratives from disjointed turns to preserve critical cross-turn dependencies.

\begin{table*}[htbp]
  \centering
  \resizebox{\textwidth}{!}{
  \renewcommand{\arraystretch}{0.92}
  \begin{tabular}{lccccccc}
    \toprule
    \multirow{2}{*}{\textbf{Method}} & \textbf{Temporal} & \textbf{Multi-Session} & \textbf{Knowledge-Update} & \textbf{Single-User} & \textbf{Single-Assistant} & \textbf{Single-Preference} & \textbf{Overall} \\
    & (n=133) & (n=133) & (n=78) & (n=70) & (n=56) & (n=30) & (N=500) \\
    \midrule
    {Full Text} & 31.58 & 45.45 & 76.92 & 87.14 & 89.29 & 36.67 & 56.89 \\
    {Naive RAG} & 39.85 & 48.48 & 67.95 & \underline{90.00} & \textbf{98.21} & 53.33 & 60.90 \\
    {LangMem}   & 15.79 & 20.30 & 66.67 & 60.00 & 46.43 & 60.00 & 37.20 \\
    {A-Mem}     & 47.36 & 48.87 & 64.11 & \textbf{92.86} & \underline{96.43} & 46.67 & 62.20 \\
    {MemoryOS}  & 32.33 & 31.06 & 48.72 & 80.00 & 64.29 & 30.00 & 44.66 \\
    {Mem0}      & 40.15 & 46.21 & 70.12 & 81.43 & 41.07 & 60.00 & 53.51 \\
    {LightMem}  & \textbf{67.18} & \textbf{71.74} & \textbf{83.12} & 87.14 & 32.14 & \underline{68.18} & \underline{69.81} \\
    \textbf{ES-Mem (Ours)} & \underline{64.66} & \underline{66.17} & \underline{78.21} & 84.29 & 82.14 & \textbf{73.33} & \textbf{72.40} \\
    \bottomrule
  \end{tabular}%
  }
  \caption{Performance comparison on LongMemEval-S benchmark using GPT-4o-mini. We use the LLM Judge Score to evaluate answer accuracy. The best performance is marked in bold, and the second best is underlined.}
  \label{tab:longmem-results}
\end{table*}

\paragraph{Results on LongMemEval-S.}
The evaluation results on the LongMemEval-S benchmark are presented in Table \ref{tab:longmem-results}. ES-Mem achieves superior performance, securing the top rank with an Overall score of 72.40 and surpassing the strong competitor LightMem. 
Most notably, ES-Mem achieves the highest accuracy score of 73.33 in the Single-Preference category, demonstrating that our fine-grained event boundaries effectively preserve specific user constraints against context dilution.
Furthermore, ES-Mem maintains robust competitiveness in Temporal, Multi-Session, and Knowledge-Update tasks, consistently ranking second. 
This adaptability originates from the structural precision of our framework, which encapsulates conflicting or time-variant information into discrete, semantically homogeneous event units and mitigates interference between historical states and valid updates during the retrieval process.
Additionally, we provide a case study of ES-Mem in Appendix \ref{sec:case_study}.

\subsection{Efficiency Analysis}
Table \ref{tab:efficiency-performance} compares efficiency in terms of token consumption and response latency.
While LangMem achieves the lowest token consumption of 127, it is significantly hampered by an excessive latency of 18.53s. 
Baselines such as Mem0 and LightMem demonstrate high operational efficiency with low latencies of 0.708s and 1.838s, respectively. Nevertheless, they still fall short of our method in terms of comprehensive generation quality. 
ES-Mem strikes an optimal balance between cost and performance. It maintains a moderate token budget of 2,925 and a low latency of 1.423s, while achieving the best performance with an Overall F1 of 45.56 and BLEU-1 of 39.70. 
This demonstrates that ES-Mem delivers superior retrieval precision with negligible computational overhead. 
For further experimental analysis investigating how the number of retrieved memory units impacts performance, please refer to Appendix \ref{sec:topk}.

\begin{table}[htbp]
\centering
\resizebox{\columnwidth}{!}{
\renewcommand{\arraystretch}{0.93}
\begin{tabular}{lcccc}
\toprule
\textbf{Method} & \textbf{Tokens} & \textbf{Latency (s)} & \textbf{Overall F1} & \textbf{Overall B1} \\
\midrule
A-Mem    & 2712 & 1.410 & 39.65 & 32.30 \\
MemoryOS & 3874 & 8.657 & 42.84 & 35.53 \\
Zep      & 3911 & 1.292 & 43.56 & 33.64 \\
Mem0     & 1764 & \textbf{0.708} & 45.09 & 35.91 \\
LangMem  & \textbf{127}  & 18.53 & 36.87 & 30.06 \\
Nemori   & 4767 & 2.512 & 44.72 & 36.75 \\
LightMem   & 815 & 1.838 & 44.73 & 37.74 \\
\textbf{ES-Mem (Ours)} & 2925 & 1.423 & \textbf{45.56} & \textbf{39.70} \\
\bottomrule
\end{tabular}
}
\caption{Efficiency analysis on LoCoMo benchmark. Reported metrics include token consumption and latency during the retrieval and generation phases.}
\label{tab:efficiency-performance}
\end{table}

\begin{figure*}[t]
    \centering
    \includegraphics[width=\linewidth]{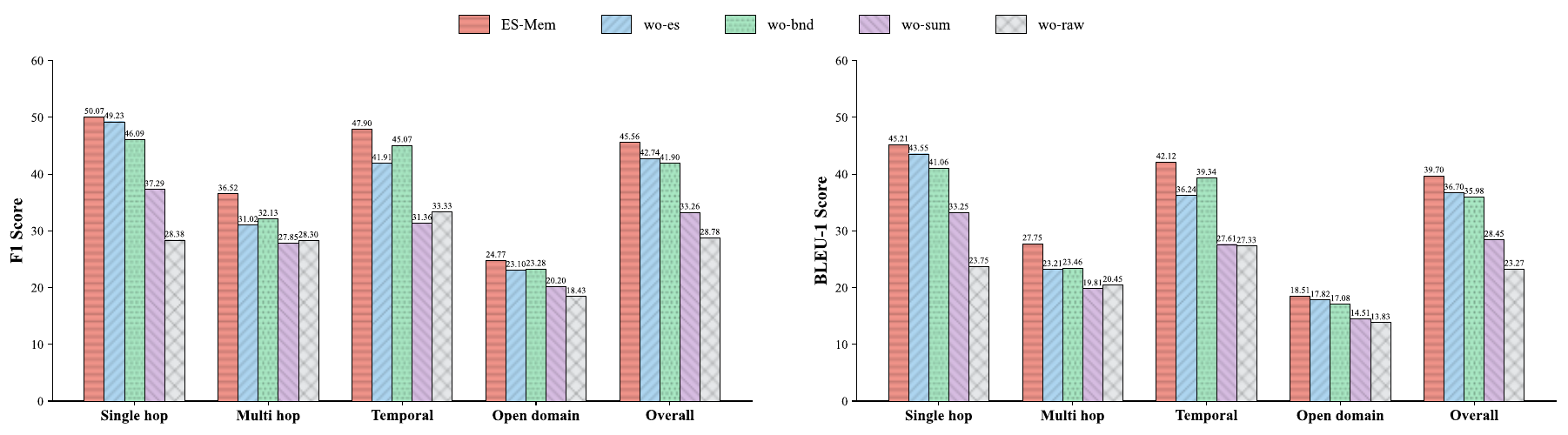}
    \caption{Results of ablation experiments on LoCoMo benchmark. We use GPT-4o-mini as the backbone LLM.}
    \label{fig:ablation}
\end{figure*}

\subsection{Ablation Study}
To validate the contributions of ES-Mem's components, we conduct ablation studies on the LoCoMo benchmark using GPT-4o-mini. There are four variants: \textbf{w/o es} (removing event segmentation), \textbf{w/o bnd} (removing boundary anchors), \textbf{w/o sum} (removing summaries), and \textbf{w/o raw} (removing raw context).
As shown in Figure \ref{fig:ablation}, the full ES-Mem framework consistently outperforms all variants, confirming the necessity of each module.
Specifically, w/o es exhibits significant degradation in Multi Hop and Temporal tasks, validating that dynamic segmentation preserves the semantic integrity required for complex reasoning.
The removal of boundaries notably impairs Multi Hop performance, proving that cognitive anchors are essential for reconstructing long-range narratives.
Conversely, w/o sum leads to a sharp decline in Single Hop tasks, highlighting the role of summaries as high-fidelity indices.
Finally, w/o raw suffers the most severe overall drop, demonstrating that raw context is indispensable for generating detail-rich responses.

\begin{table*}[htbp]
\centering
\resizebox{\textwidth}{!}{%
\setlength{\tabcolsep}{9pt} 
\renewcommand{\arraystretch}{0.79}
\begin{tabular}{lcccccccccccc}
\toprule
\multirow{2}{*}{Method} & \multicolumn{4}{c}{DialSeg711} & \multicolumn{4}{c}{SuperDialSeg} & \multicolumn{4}{c}{TIAGE} \\
\cmidrule(lr){2-5} \cmidrule(lr){6-9} \cmidrule(lr){10-13}
 & Pk$\downarrow$ & WD$\downarrow$ & F1$\uparrow$ & Score$\uparrow$ & Pk$\downarrow$ & WD$\downarrow$ & F1$\uparrow$ & Score$\uparrow$ & Pk$\downarrow$ & WD$\downarrow$ & F1$\uparrow$ & Score$\uparrow$ \\
\midrule
\multicolumn{13}{c}{Unsupervised Methods} \\
\midrule
BayesSeg & 0.306 & 0.350 & 0.556 & 0.614 & 0.433 & 0.593 & 0.438 & 0.463 & 0.486 & 0.571 & 0.366 & 0.419 \\
TextTiling & 0.470 & 0.493 & 0.245 & 0.382 & 0.441 & 0.453 & 0.388 & 0.471 & 0.469 & 0.488 & 0.204 & 0.363 \\
GraphSeg & 0.412 & 0.442 & 0.392 & 0.483 & 0.450 & 0.454 & 0.249 & 0.398 & 0.496 & 0.515 & 0.238 & 0.366 \\
GreedySeg & 0.381 & 0.410 & 0.445 & 0.525 & 0.490 & 0.494 & 0.365 & 0.437 & 0.490 & 0.506 & 0.181 & 0.341 \\
\midrule
\multicolumn{13}{c}{Supervised Methods} \\
\midrule
BERT & - & - & - & - & 0.214 & 0.225 & 0.725 & 0.745 & 0.418 & 0.435 & 0.124 & 0.349 \\
RoBERTa & - & - & - & - & 0.185 & 0.192 & 0.784 & 0.798 & 0.265 & 0.287 & 0.572 & 0.648 \\
RetroTS-T5 & - & - & - & - & 0.227 & 0.237 & 0.733 & 0.750 & 0.280 & 0.317 & 0.576 & 0.639 \\
\midrule
\multicolumn{13}{c}{LLM-based Methods} \\
\midrule
Def-DTS & 0.500 & 0.380 & 0.133 & 0.346 & 0.480 & 0.518 & 0.179 & 0.340 & 0.455 & 0.282 & 0.420 & 0.526 \\
SeCom & 0.447 & 0.316 & 0.181 & 0.400 & 0.492 & 0.391 & 0.316 & 0.437 & 0.523 & 0.408 & 0.397 & 0.465 \\
\textbf{ES-Mem (Ours)} & 0.172 & 0.098 & 0.692 & 0.778 & 0.434 & 0.283 & 0.411 & 0.526 & 0.382 & 0.268 & 0.556 & 0.616 \\
\bottomrule
\end{tabular}
}
\caption{Dialogue segmentation performance on three datasets.  Unsupervised methods are reported in \citet{superdialseg}. Supervised methods are reported in \citet{def-dts}. No supervised learning results for Dialseg711 due to missing training/validation split.}
\label{tab:segmentation_results}
\end{table*}

\subsection{Evaluation of Event Segmentation Module}\label{4.6}
To assess the robustness of our event segmentation module independently, particularly when deployed with resource-constrained backbones, we utilize Qwen2.5-3B as the underlying LLM and conduct extensive evaluations on three widely used dialogue segmentation datasets: DialSeg711 \citep{dilseg711}, SuperDialSeg \citep{superdialseg}, and TIAGE \citep{tiage}.
For evaluation metrics, we follow \citep{superdialseg} to employ F1, $\mathrm{P_k}$ \citep{pk}, WindowDiff (WD) \citep{wd} and a composite segmentation score\footnote{ \url{https://2023.ieeeicassp.org/signal-processing-grand-challenges/}}:
\begin{equation}
    \mathrm{Score} = \frac{2 \cdot \mathrm{F1} + (1 - \mathrm{P_k}) + (1 - \mathrm{WD})}{4}
\end{equation}

As shown in Table \ref{tab:segmentation_results}, our proposed method achieves significant performance gains across all three datasets. 
It consistently outperforms existing unsupervised approaches and other LLM-based baselines, specifically Def-DTS \citep{def-dts} and SeCom \citep{secom}. Furthermore, on the TIAGE dataset, our method even surpasses several supervised models, demonstrating its adaptability in capturing semantic transitions without relying on extensive domain-specific supervision.

\section{Conclusion}
In this work, we propose ES-Mem to address the limitations of existing memory frameworks, specifically regarding rigid memory granularity and the lack of structural guidance in flat retrieval paradigms.
By innovatively integrating Event Segmentation Theory into dialogue agents, we design a dynamic event segmentation module and a hierarchical memory architecture to emulate human cognitive processes. ES-Mem partitions long-term dialogues into semantically homogeneous events based on topical and intent coherence. The resulting refined event boundaries serve as high-level cognitive anchors, facilitating a coarse-to-fine retrieval strategy. This mechanism precisely locates relevant memory intervals, effectively retrieving coherent episodic contexts from massive dialogue history.
We conduct systematic experiments on two long-term memory benchmarks. The results demonstrate that ES-Mem significantly outperforms baseline methods. Additionally, our event segmentation module exhibits robust adaptability on multiple dialogue segmentation datasets.

\section*{Limitations}
Although ES-Mem demonstrates significant improvements in long-term memory performance, several limitations remain to be addressed in future research.
First, our current framework has not fully modeled the dynamic evolution of memory. 
While ES-Mem effectively segments and stores episodic events, the stored memories currently remain relatively static. 
We have not yet implemented mechanisms that mimic complex human cognitive processes such as memory consolidation, the natural decay of non-salient details, or the abstraction of repeated episodes into higher-level schemas over time. 
Future work will focus on introducing dynamic evolution mechanisms to better simulate the plasticity and reconstructive nature of human memory.
Moreover, the current system is primarily designed for text-modal interaction data. 
Future research can explore how to extend the event segmentation and hierarchical retrieval paradigms to accommodate diverse input modalities, such as images, audio, and video. 
Incorporating multimodal data would provide agents with richer forms of memory storage and significantly enhance their adaptability in real-world interactive scenarios.

\section*{Ethics Statement}
This work introduces a memory framework for dialogue agents. We utilize publicly available datasets (LoCoMo, LongMemEval, etc.) that are widely used in the research community. We do not collect any private user data or personally identifiable information. The proposed method is intended to enhance the coherence of dialogue agents and does not inherently introduce harmful biases or generate toxic content beyond the capabilities of the underlying base LLMs. This work complies with the ACL Ethics Policy.

\bibliography{custom}

\clearpage
\appendix
\section{Case Study}
\label{sec:case_study}
\begin{figure*}[t]
    \centering
    \includegraphics[width=\textwidth]{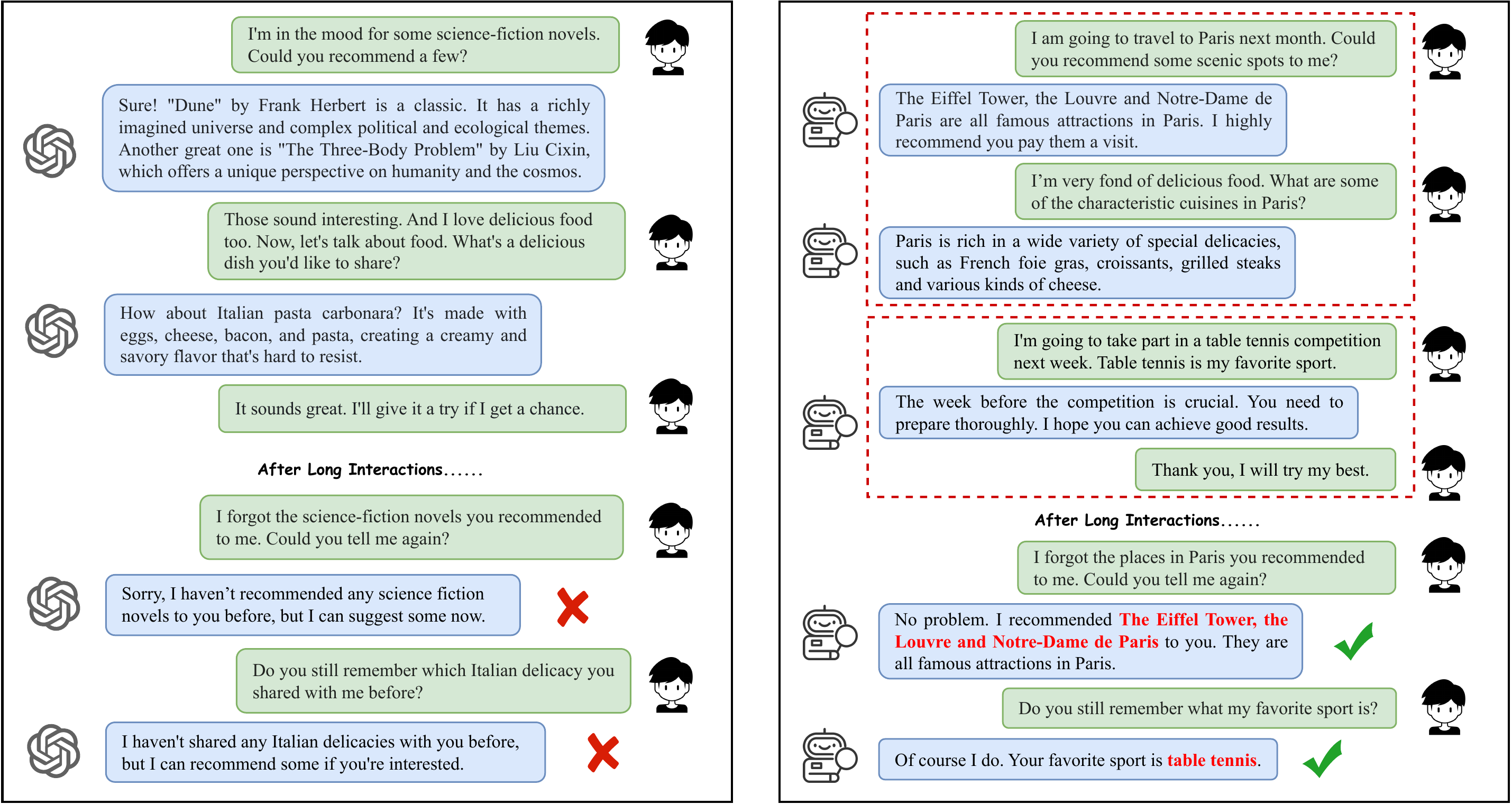}
    \caption{Case study in long-term dialogue scenarios.}
    \label{fig:case}
\end{figure*}

Figure \ref{fig:case} presents a comparative qualitative analysis between a standard LLM (left) and the proposed ES-Mem framework (right) in a long-term dialogue setting.
The left panel illustrates the limitations of a standard LLM lacking a dedicated memory module: after extended interactions, the model suffers from context loss, failing to recall specific prior recommendations (e.g., \textit{Sci-Fi novels} and \textit{Italian delicacies}) and producing hallucinatory responses.
In contrast, the right panel demonstrates the efficacy of ES-Mem. Our system dynamically partitions the continuous dialogue stream into semantically coherent events—highlighted as the \textit{Paris Travel} phase and the \textit{Table Tennis} discussion—thereby preserving distinct episodic contexts.
Leveraging these event boundaries as retrieval anchors, ES-Mem successfully overcomes the context amnesia problem, accurately retrieving fine-grained details (e.g., specific scenic spots and user preferences) despite the long temporal gap.
This comparison highlights ES-Mem's ability to maintain long-term consistency through its structured, event-driven memory architecture.

\section{Parameter Setup}
\label{sec:setup}
All the experiments are conducted on hardware equipped with an NVIDIA GeForce RTX 2080 Ti GPU (11 GB), an Intel Core i9-9820X CPU @ 3.30GHz, and 96 GB of RAM. We use the following hyper-parameters for our experiments:

\begin{itemize}
  \item \textbf{Dynamic Event Segmentation:} 
  \begin{itemize}
    \item Quantile $q$: 0.35
    \item Boundary Probability Threshold $\tau_c$: 0.75
    \item Context Window Length $L$: 2
  \end{itemize}
\end{itemize}

\begin{itemize}
  \item \textbf{Memory Construction and Retrieval:} 
  \begin{itemize}
    \item Refined Boundary Context Length $L$: 4
    \item Anchor Candidates $k$: 10
    \item Expansion Window Size $w$: 3
    \item Fusion Weight $\alpha$: 0.70
    \item Final Top-$K$ Events: 5 for Qwen2.5-3B/Llama3.2-3B; 10 for GPT-4o-mini
  \end{itemize}
\end{itemize}

\begin{figure*}[t]
    \centering
    \includegraphics[width=\textwidth]{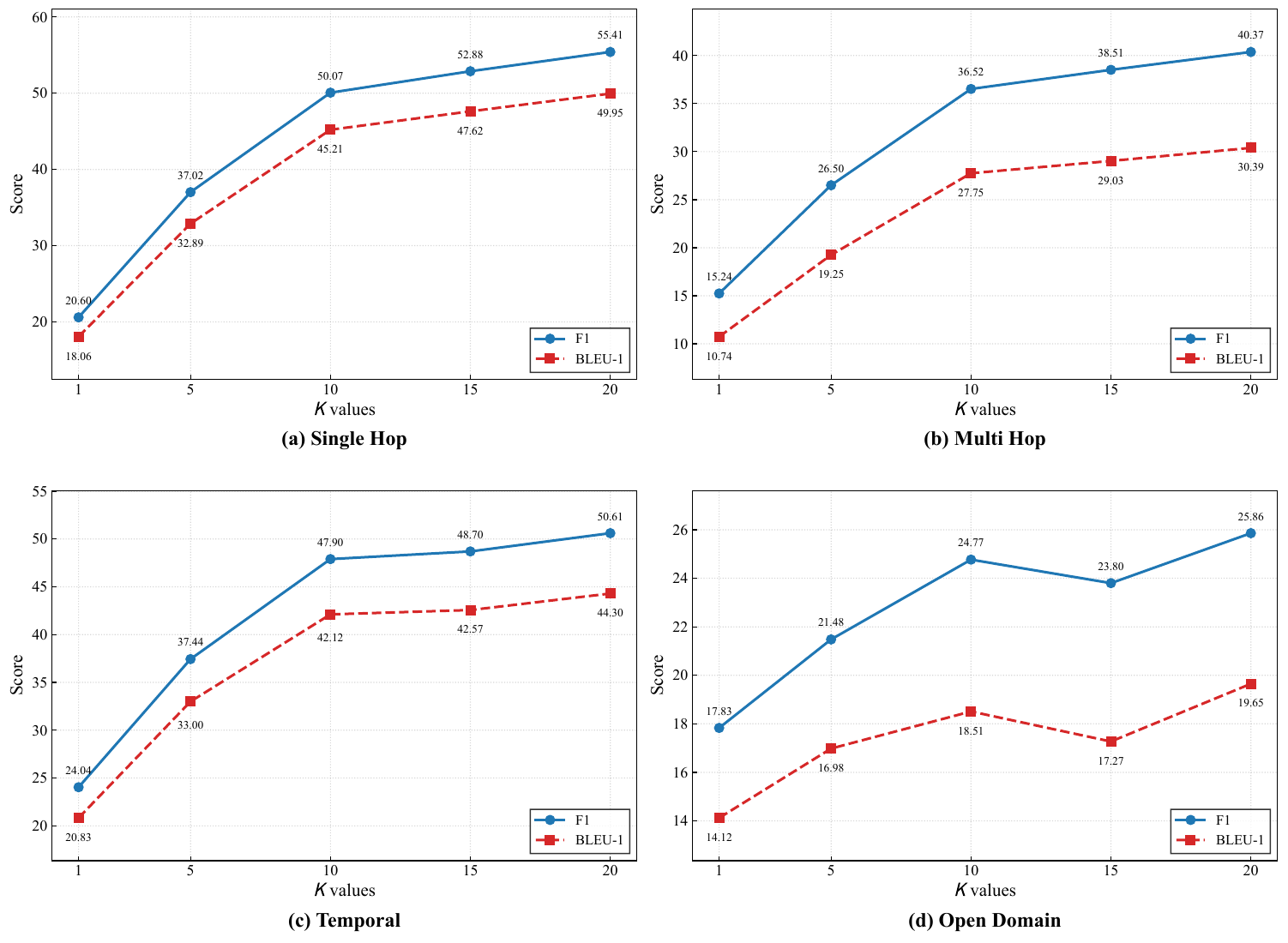}
    \caption{Results of memory retrieval parameter $K$. Using GPT-4o-mini as the base LLM model.}
    \label{fig:top-k}
\end{figure*}

\section{The Impact of Top-$K$}
\label{sec:topk}
To evaluate the impact of retrieval volume on memory performance, we conduct a sensitivity analysis on the number of retrieved events $K$ (where $K \in \{1, 5, 10, 15, 20\}$) using the GPT-4o-mini backbone on the LoCoMo benchmark.
As illustrated in Figure \ref{fig:top-k}, performance metrics exhibit a rapid upward trajectory across all tasks as $K$ increases from 1 to 10, confirming that sufficient context is a prerequisite for effective reasoning.
However, beyond the inflection point of $K=10$, marginal gains diminish significantly. In Open Domain tasks, performance even fluctuates (e.g., a slight dip at $K=15$), suggesting that excessive context introduces irrelevant noise that may distract the model.
Consequently, we identify $K=10$ as the optimal setting, striking a robust balance between maximizing information recall and minimizing noise interference and computational overhead.

\section{Details about Baselines}
\label{sec:baselines}
\paragraph{MemGPT} MemGPT \citep{MemGPT} introduces the notion of hierarchical memory management from traditional operating systems, providing scalable long-range context management for dialogue agents.
\paragraph{MemoryBank} MemoryBank \citep{Memorybank} dynamically adjusts memory strength according to the Ebbinghaus forgetting curve, and supports continual memory updating and user profile construction.
\paragraph{A-Mem} A-Mem \citep{A-mem} is an associative memory framework inspired by the Zettelkasten note-taking method. It models each memory as a note with multiple structured attributes and dynamically creates and updates links between memories, yielding an evolving knowledge network.
\paragraph{MemoryOS} MemoryOS \citep{memoryos} is a memory management framework of operating systems to build a three-tier storage architecture comprising short-term, medium-term, and long-term personalized memories. Different tiers of memory employ different manage strategies.
\paragraph{H-Mem} H-Mem \citep{h-mem} constructs a four-level hierarchical storage structure based on semantic abstraction. It employs positional index encoding for layer-wise retrieval and integrates a dynamic update mechanism based on the Ebbinghaus Forgetting Curve and user feedback.
\paragraph{Zep} Zep \citep{zep} leverages a temporal knowledge graph engine with a three-tiered subgraph structure comprising episodic, semantic entity, and community layers. It employs bi-temporal modeling for data integration and incorporates hybrid search with multi-strategy reranking mechanisms.
\paragraph{Mem0} Mem0 \citep{mem0} adopts a two-stage pipeline of memory extraction and update. It utilizes an LLM to extract salient memory candidates from conversation contexts and subsequently performs adaptive operations (addition, update, or deletion) based on semantic similarity matching.
\paragraph{LangMem} LangMem\footref{fn:langmem} integrates episodic and procedural memory types managed by a dedicated Memory Manager. It supports storage and retrieval across diverse modes, including cumulative knowledge bases, user profiles, and procedural workflows.
\paragraph{Nemori} Nemori \citep{nemori} constructs dual databases for episodic and semantic memory. It employs a two-step alignment mechanism for structural organization and a "Predict-Calibrate" principle to drive the self-organization and evolution of the memory system.
\paragraph{LightMem} LightMem \citep{lightmem} is inspired by the Atkinson-Shiffrin human memory model and balances performance and efficiency for LLMs' memory systems via cognition-inspired sensory memory, topic-aware short-term memory for summarization, and long-term memory with offline sleep-time updates.

\section{Prompt Templates}
\label{sec:prompt}
\begin{figure*}[b]
    \centering
    \includegraphics[width=\textwidth]{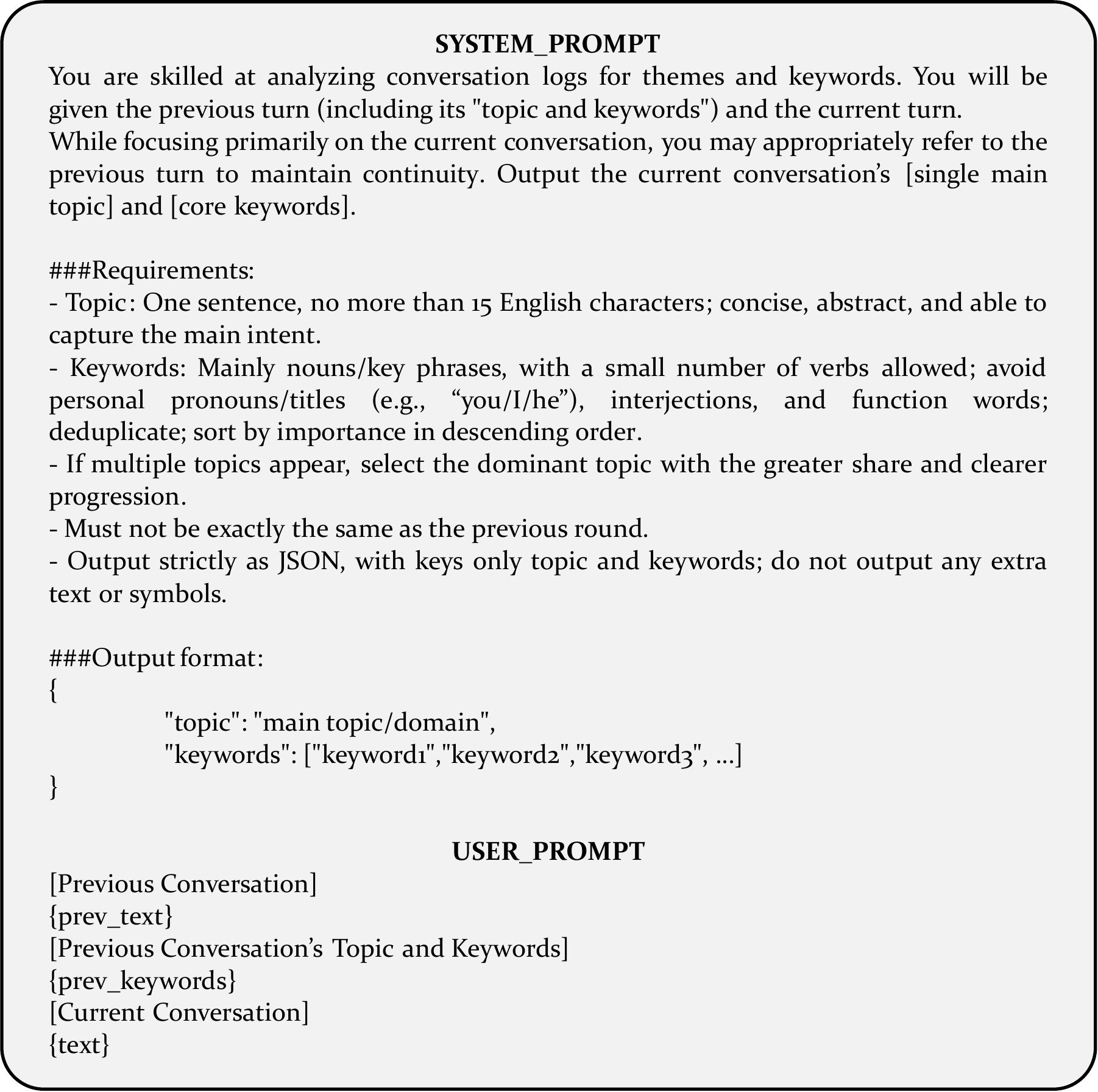}
    \caption{The prompt for extracting topic and keywords of each dialogue turn.}
    \label{tab:topic-keywords}
\end{figure*}

\begin{figure*}[t!] 
    \centering
    \includegraphics[width=\textwidth]{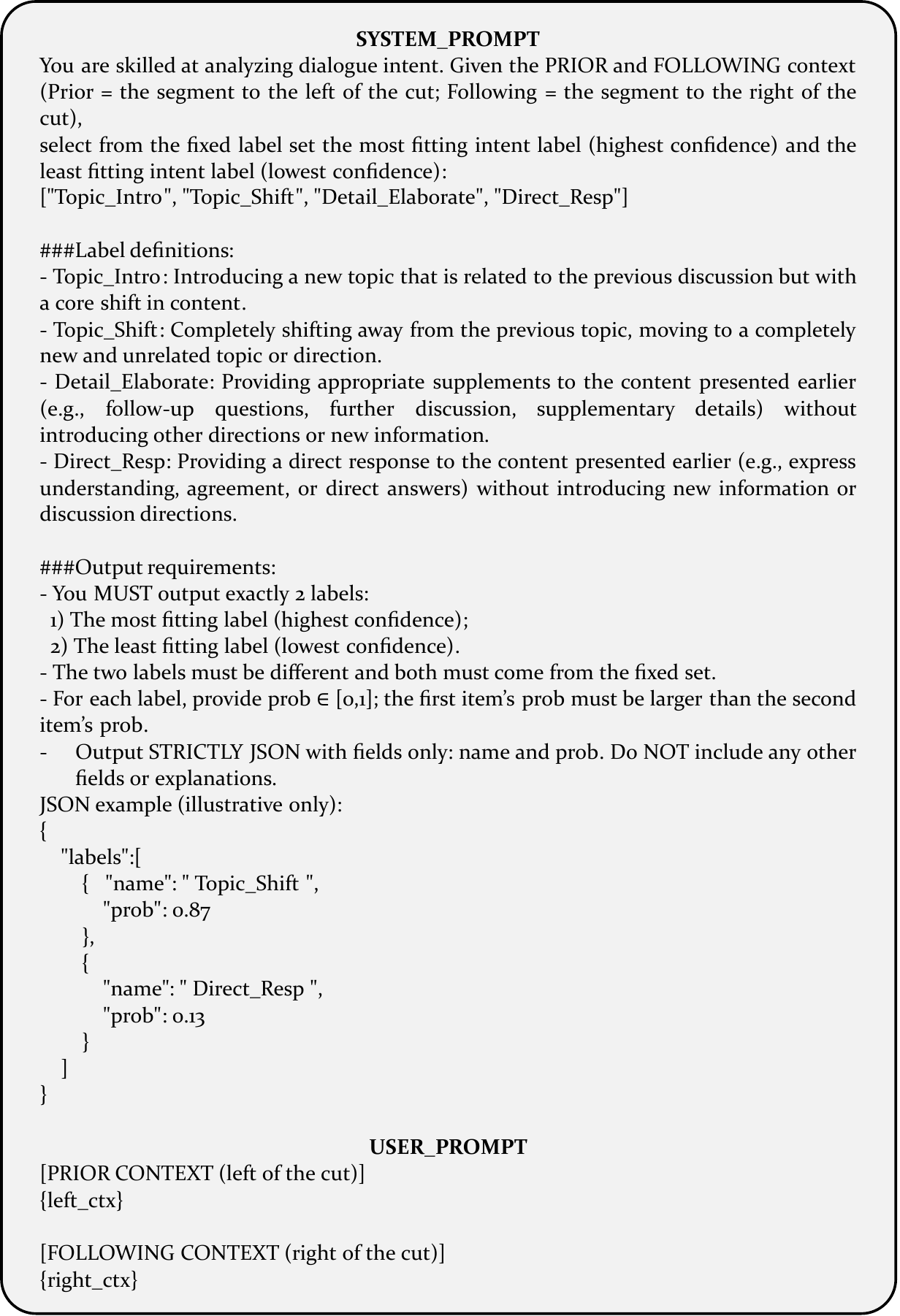}
    \caption{The prompt for recognizing intent labels.}
    \label{tab:intent-label}
\end{figure*}

\begin{figure*}[t] 
    \centering
    \includegraphics[width=\textwidth]{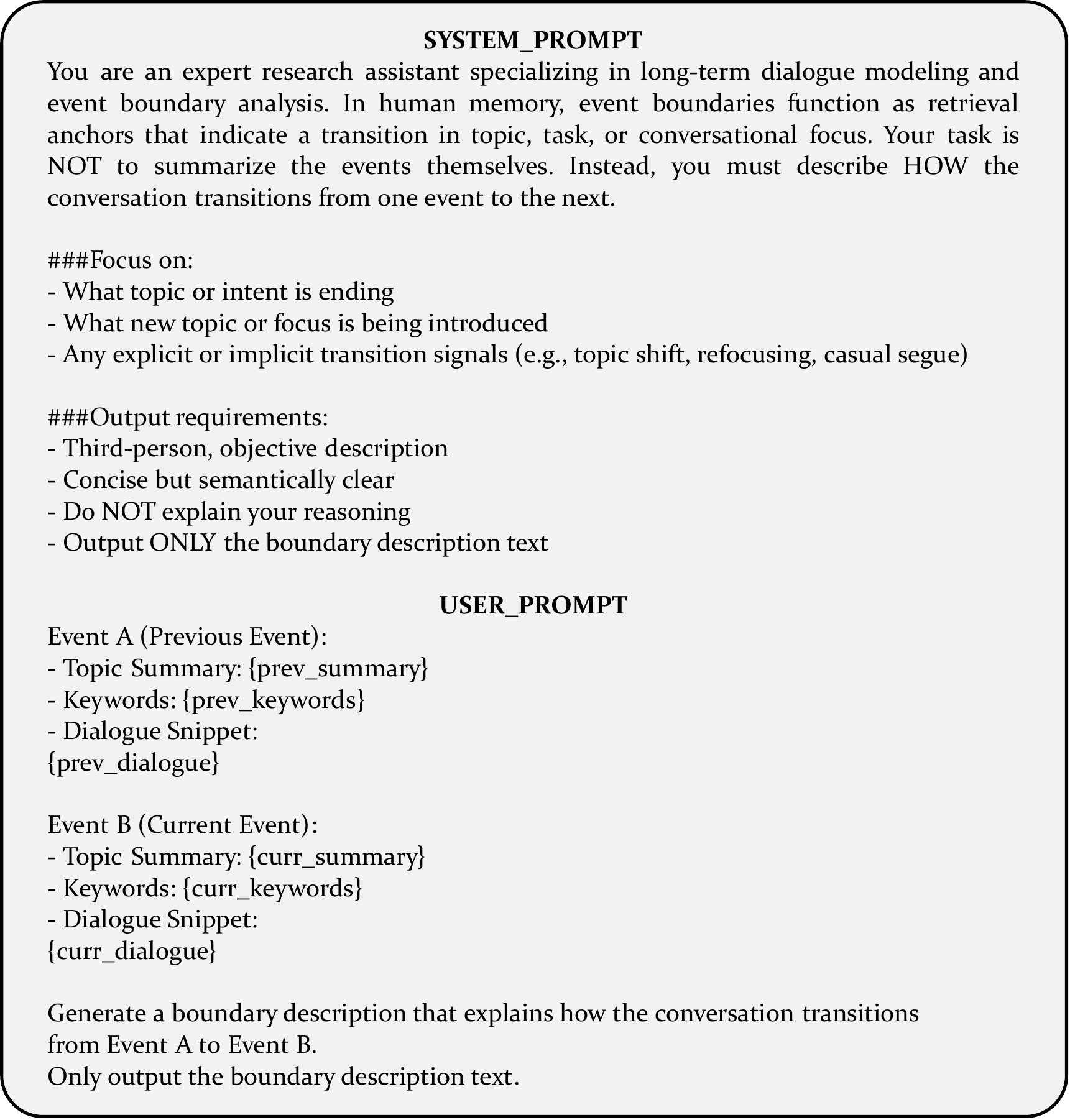}
    \caption{The prompt for generating refined event boundaries.}
    \label{tab:boundary}
\end{figure*}

Figures \ref{tab:topic-keywords}, \ref{tab:intent-label}, \ref{tab:boundary} present the core prompt templates of the ES-Mem framework.

\section{Data License}
We utilize five publicly available datasets in our experiments, ensuring full compliance with their respective licenses for non-commercial research purposes. 
LoCoMo is distributed under the Creative Commons Attribution-NonCommercial 4.0 International (CC BY-NC 4.0) license. 
LongMemEval is governed by the MIT License. 
For the dialogue segmentation tasks, we employ three standard benchmarks—DialSeg711, SuperDialSeg, TIAGE—which are open-source datasets widely used in discourse analysis research.

\end{document}